\documentclass[runningheads]{llncs}
\usepackage{graphicx}

\usepackage{tikz}
\usepackage{comment}
\usepackage{amsmath,amssymb} 
\usepackage{color}

\usepackage[noend]{algpseudocode}
\usepackage{algorithmicx,algorithm}

\usepackage[accsupp]{axessibility}  

\usepackage[colorlinks,linkcolor=red,anchorcolor=blue,citecolor=green]{hyperref}
\usepackage{url}
 
\begin{document}
\pagestyle{headings}
\mainmatter

\title{DH-AUG: DH Forward Kinematics Model Driven Augmentation for 3D Human Pose Estimation} 


\titlerunning{DH FK Model Driven Augmentation for 3D Human Pose Estimation}
%

\author{
Linzhi Huang \and
Jiahao Liang \and
Weihong Deng\textsuperscript{*}}

\authorrunning{L. Huang and W. Deng et al.}
%
\institute{Beijing University of Posts and Telecommunications \\
\email{\{huanglinzhi, jiahao.liang, whdeng\}@bupt.edu.cn}}
\maketitle

\begin{abstract}
Due to the lack of diversity of datasets, the generalization ability of the pose estimator is poor. 
To solve this problem, we propose a pose augmentation solution via DH forward kinematics model, which we call DH-AUG. 
We observe that the previous work is all based on single-frame pose augmentation, if it is directly applied to video pose estimator, there will be several previously ignored problems: (i) angle ambiguity in bone rotation (multiple solutions); (ii) the generated skeleton video lacks movement continuity. 
To solve these problems, we propose a special generator based on DH forward kinematics model, which is called DH-generator. 
Extensive experiments demonstrate that DH-AUG can greatly increase the generalization ability of the video pose estimator. 
In addition, when applied to a single-frame 3D pose estimator, our method outperforms the previous best pose augmentation method.
The source code has been released at \url{https://github.com/hlz0606/DH-AUG-DH-Forward-Kinematics-Model-Driven-Augmentation-for-3D-Human-Pose-Estimation}.

\end{abstract}

\section{Introduction}

\begin{figure}[t]
\begin{center}
\includegraphics[width=0.95\linewidth]{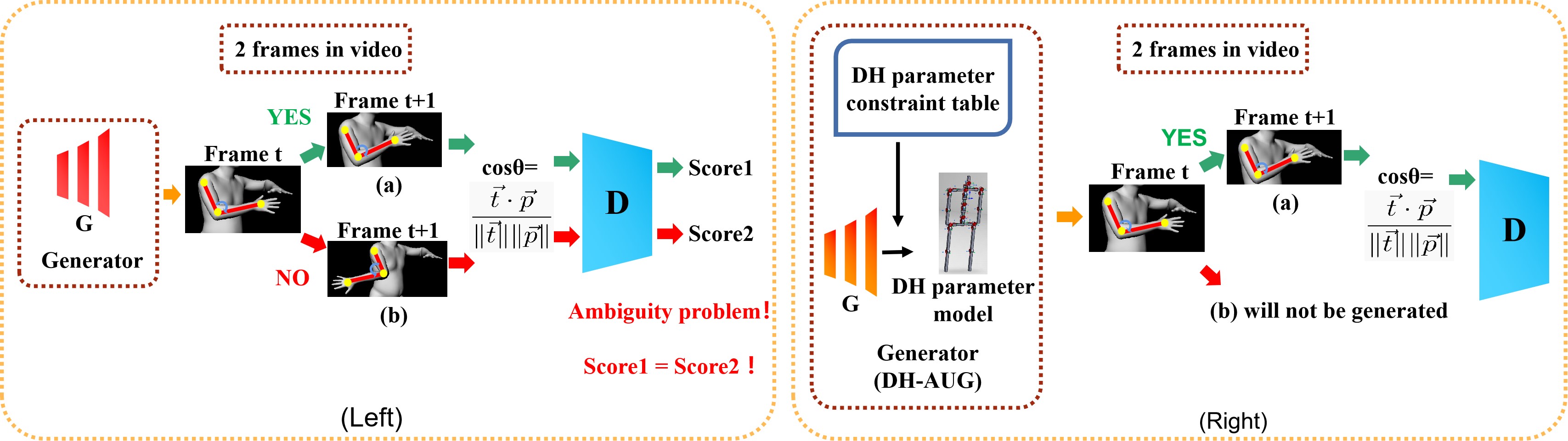}
\end{center}
\caption{\textbf{Angle ambiguity (multiple solutions)}. 
\textbf{Left}: Pose augmentation of ordinary GAN framework.
\textbf{Right}: DH-AUG.
\textbf{(a)}: Elbow rotates normally (angle is about 90°). 
\textbf{(b)}: Elbow rotates abnormally (angle is about -90°). 
Although the rotation directions of (a) and (b) are different, the cosine angle values calculated by vector inner product are the same. 
Both of them will make the discriminator output the same score, resulting in ambiguity. To weaken the ambiguity problem, we improve the generator by adding DH forward kinematics model (DH parameter model) and constraints. ${t}$ and ${p}$ are a pair of adjacent bone vectors. SMPL~\cite{loper2015smpl} is only used for visualization.
}
\label{fig:angle_ambiguity}
\end{figure}

3D pose estimation is the task of estimating 3D human pose from images.
It is a fundamental task in action recognition~\cite{li2019actional,zhang2020context,shi2019two,li2019spatio},
human tracking~\cite{mehta2017vnect}, etc.
It is difficult to obtain a 3D label, so the existing 3D data is very limited and the diversity is seriously insufficient.
This also leads to poor generalization ability of the 2D-to-3D model.

Recently, a work~\cite{li2020cascaded} enhanced data by randomly exchanging limbs, locally rotating limbs, and randomly changing bone length.
This method is dependent on the random seed, and the result is unstable.
PoseAug~\cite{gong2021poseaug} uses GAN~\cite{goodfellow2014generative} to solve the above problems.
However, PoseAug is also designed for a single-frame 3D pose estimator.
There are some problems that can not be ignored in pose augmentation in video 3D human pose estimation: angle ambiguity and angle continuity.
Most pose discriminators \cite{cheng20203d,wandt2019repnet,gong2021poseaug} calculates the cosine angle value through the inner product of two bone vectors for constraint. 
But this is a problem of multiple solutions (angle ambiguity) as shown in Fig.~\ref{fig:angle_ambiguity} (Left). 
The value calculated by the inner product corresponds to multiple angles.
For example, 0 corresponds to 90° and -90°, which makes the discriminator unable to distinguish between elbow 90° internal rotation and 90° external rotation. 
Both of them will make the discriminator output the same score, and the data distribution of the generator will contain the angle value of abnormal rotation.  
What's more, there will be discontinuous actions in the skeleton video because of angle ambiguity.
So it is not enough to use the discriminator for constraints.
We try to modify the generator to weaken this problem.
Specifically, we use DH parameters to build a human kinematics model (DH parameter model). 
This model allows us to obtain a new pose directly by changing the joint angle, and we can easily constrain the rotation direction of the joint.
We introduce this model into the generator and constrain the DH parameters so that the generator will not produce an unreasonable pose as shown in Fig.~\ref{fig:angle_ambiguity} (Right). Inspired by some previous work \cite{cheng20203d,shi2020motionet,tripathi2020posenet3d}, we also add timing information to the discriminator to increase the continuity of the generated skeleton video.

Our contributions are as follows:
\begin{itemize}
	\setlength\itemsep{-0.3em}
	\item We propose DH-AUG: a pose augmentation framework for 3D human pose estimation. It consists of DH-Generator, DH parameter model, single-frame pose discriminator and multi-stream motion discriminator.
	\item We use DH parameters to design a human kinematics model, called DH parameter model. By adding DH parameter model and constraints to the generator, the angle ambiguity is successfully weakened and the possibility of generating unreasonable pose is reduced. 
	\item 
	Extensive experiments demonstrate that DH-AUG can greatly increase the generalization ability of the video pose estimator. In addition, when applied to a single-frame 3D pose estimator, our method outperforms the previous best pose augmentation method. 
	\item 
	We release a new dataset (DH-3DP) synthesized with DH-AUG, which can be used in the 2D-to-3D network.
\end{itemize}

\section{Related Work}
\textbf{3D human pose estimation.}
There are two mainstream monocular 3D human pose estimation methods, one is to obtain 3D pose end-to-end~\cite{pavlakos2017coarse,tekin2016structured,tekin2016direct}, and the other is through the multi-stage method, first obtain 2D pose from the images~\cite{su2019cascade,chu2017multi,tang2018deeply}, and then further obtain 3D pose from 2D pose~\cite{martinez2017simple,li2019generating,xu2020deep,cheng20203d}.
The second method is more common.
We do not pay too much attention to the model structure.  
We focus on pose augmentation for 2D-to-3D networks and produce 2D-3D pairs.
According to the input mode, it can be divided into single-frame input and video input. 
Video input can weaken the depth ambiguity problem~\cite{pavllo20193d,zheng20213d,cai2019exploiting}.
We design a pose augmentation scheme for single-frame pose estimation and video pose estimation.

\textbf{Kinematic model.} 
The kinematic model is widely used in the field of the robot~\cite{csiszar2017solving}, hand pose estimation~\cite{mueller2017real,kokic2019learning}, and games. Recent work~\cite{li2021hybrik} uses forward and inverse kinematics to make up for the shortcomings of 3D pose estimation and mesh parameter models. 
Inspired by this, we use the DH parameter~\cite{craig2009introduction} to build a 3D human forward kinematics model to weaken the angle ambiguity.
DH parameter is a method to describe the coordinate system of connecting links.

\textbf{pose augmentation for 3D human pose estimation.} 
Due to the high cost of 3D data acquisition and insufficient data diversity,
the 2D-to-3D model is difficult to have good generalization ability. 
In some works, pose augmentation of 3D pose estimation is carried out by synthesizing images~\cite{peng2018jointly,rogez2016mocap,varol2017learning}.
It is worth noting that there is another way to obtain new data pairs by synthesizing 2D and 3D data. 
The recently proposed evolutionary algorithm~\cite{li2020cascaded} uses random exchange, local rotation to generate data.
The data generated in this way has great randomness, depending on the preset parameters.
PoseAug\cite{gong2021poseaug} proposes to use GAN with a feedback mechanism to generate data, which is more effective than the former.
However, this method has insufficient constraints on joint rotation.
This is not conducive to being extended to video pose estimation. 
Therefore, we propose a combination of the DH parameter model and GAN for pose augmentation.

\section{Method}

\subsection{Overview}
\begin{figure}[t]
\begin{center}
\includegraphics[width=0.95\linewidth]{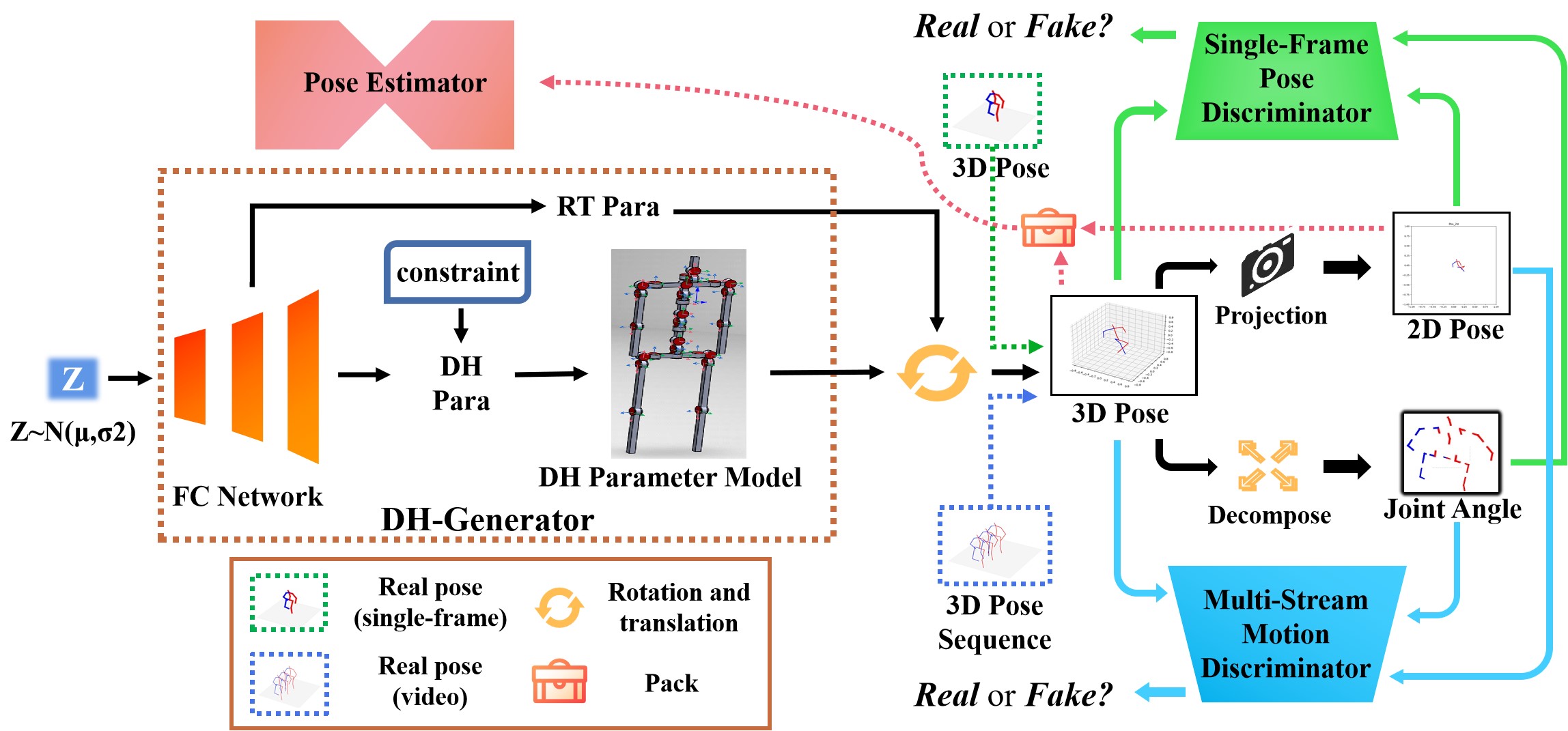}
\end{center}
\caption{\textbf{Overview of the overall framework of DH-AUG.} 
128-dimensional vectors are sampled from the normal distribution and input into the fully connected network to obtain DH parameters, global rotation and translation parameters. 
Then, the 3D pose is obtained through DH parameter model.
\textbf{1) Single-frame}: The 3D pose, 2D pose and joint angle are transmitted to the single-frame pose discriminator for training.
\textbf{2) Video}: Input 3D pose sequence, 2D pose sequence, bone rotation trajectory (joint angle) into single-frame pose discriminator
and multi-stream motion discriminator. Finally, the newly generated 2D-3D data pair is packaged
into a new dataset and transmitted to the pose estimator for training.}
   \label{fig:framework_dh_aug}
\end{figure}

There are multiple solutions for mapping the coordinates of 3D keypoints to the angle value, so it is not enough to use the discriminator for constraints. To weaken the angle ambiguity problem and further improve the effect of pose augmentation, we introduce DH parameters into GAN framework, as shown in Fig.~\ref{fig:framework_dh_aug}. 
We use the fully connected network to generate DH parameters, etc., and transfer them into the DH parameter model to obtain the corresponding 3D pose. In addition, we also use discriminators to force the generator to generate more reasonable and diversified 3D pose. It is worth noting that we add constraints to the DH parameter model to avoid generating unreasonable pose and weaken the angle ambiguity. More specific contents will be introduced in this section.

\begin{figure}[t]
\begin{center}
\includegraphics[width=0.95\linewidth]{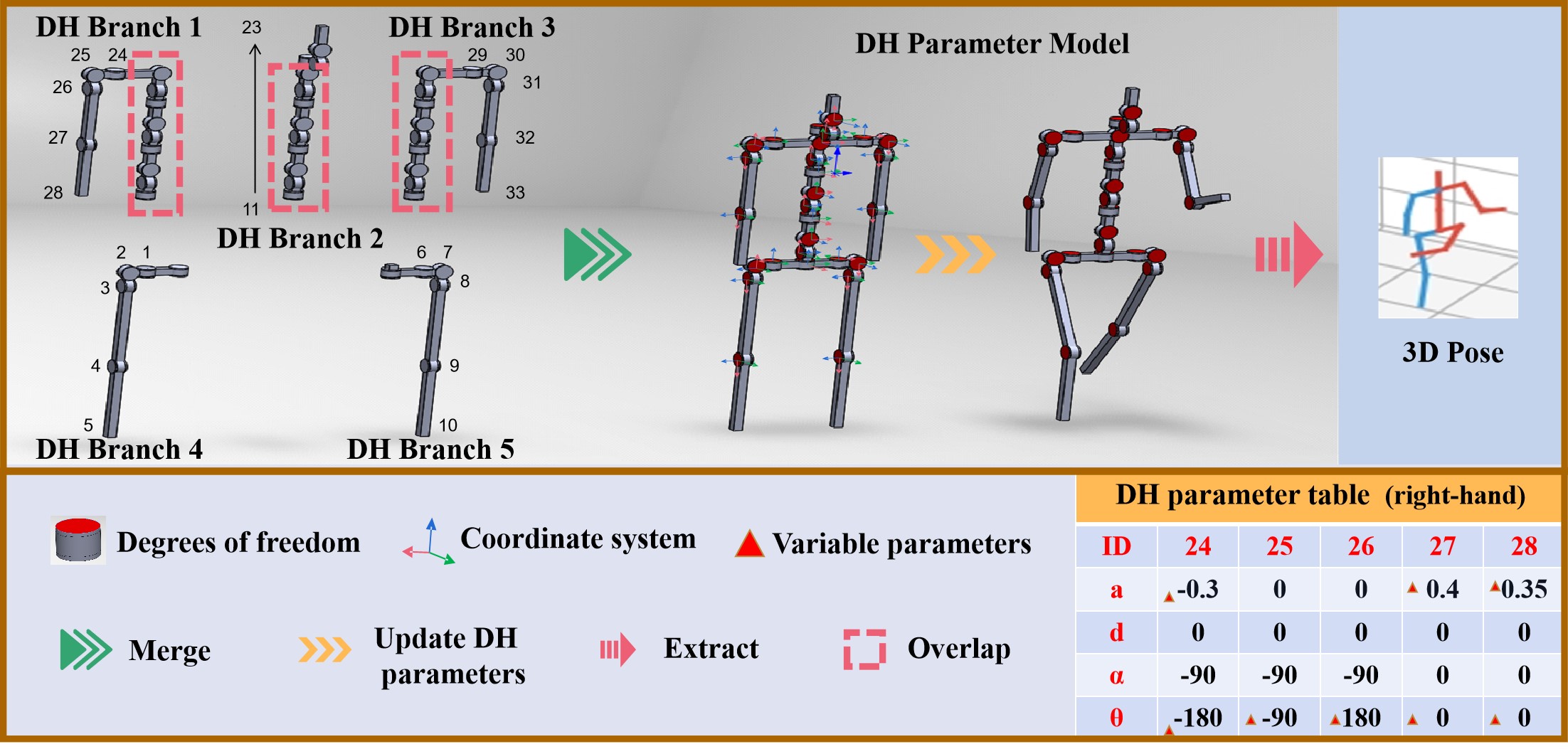}
\end{center}
\caption{\textbf{DH parameter model.} 
There are 33 degrees of freedom (DOF) and 48 changeable DH parameters. 5 DOF in the figure are not drawn (head, ankle, and wrist). 
We built 5 DH branches. 
The root node is the hip, and the overlapping parts share DH parameters. 
The part sharing DH parameters combines 5 branches into a complete human kinematics model. 
Then the new transformation matrix is obtained by updating the DH parameters. 
Finally, a new 3D pose is extracted from the transformation matrix. 
(See Alg.~\ref{alg1} for the process of building a human kinematics model with DH parameters. 
The complete DH parameter table is in the supplementary material.)}
   \label{fig:dh_para_model}
\end{figure}

\subsection{DH Parameter Model}
\begin{figure}[t]
\begin{center}
\includegraphics[width=0.95\linewidth]{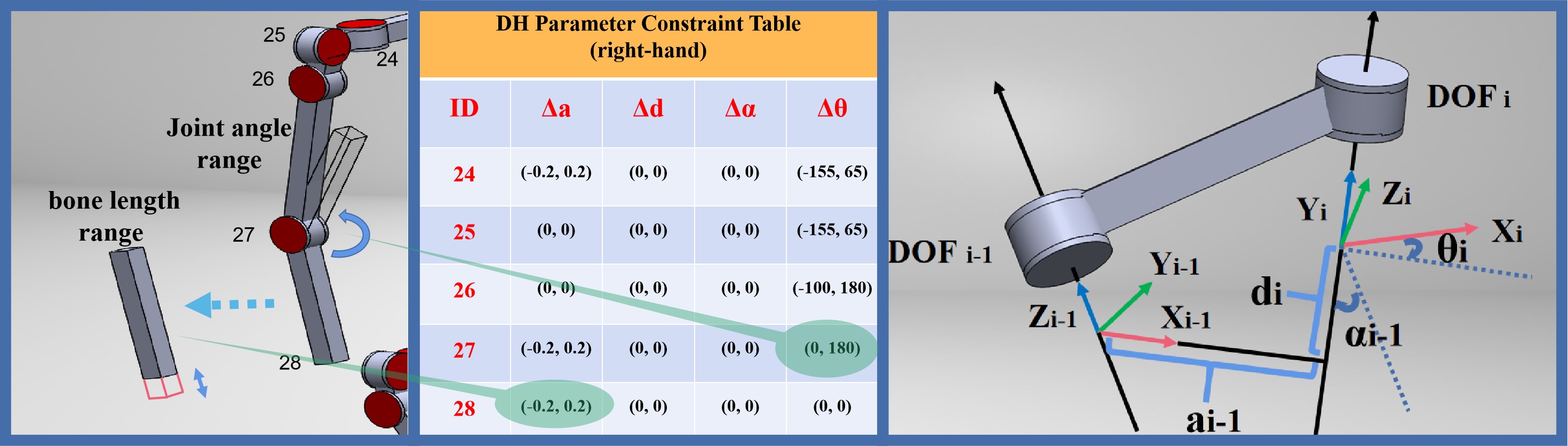}
\end{center}
\caption{
\textbf{Left}: The constraint diagram of the elbow. 
\textbf{Middle}: DH parameter constraint table.
\textbf{Right}: The schematic diagram of DH parameter~\cite{craig2009introduction}. 
$a$ is link length, $d$ is link offset, $\alpha$ is twist angle, $\theta$ is the joint angle.
$\Delta a, \Delta d, \Delta \alpha, \Delta \theta$ is the change in the DH parameter.
\textbf{(The complete DH parameter constraint table can be seen in the supplementary material.)}}
   \label{fig:right hand}
\end{figure}

\begin{algorithm}[]
\caption{DH parameter model}\label{alg1}
\textbf{Input:}$\Delta a, \Delta d, \Delta \alpha, \Delta \theta, R_x, R_y, R_z, T_x, T_y, T_z$\\
\textbf{Output:}$P_{new}$

\begin{algorithmic}
\For{$i$ \textbf{in} $N_{branch}$}
    \begin{algorithmic}
    \For{$k$ \textbf{in} $N_{Dof(i)}$}
        \State
        $A$ = $a_{i,k}$+$\Delta a_{i,k}$; 
        $B$ = $d_{i,k}$+$\Delta d_{i,k}$;
        $C$ = $\alpha_{i,k}$+$\Delta \alpha_{i,k}$; 
        $D$ = $\theta_{i,k}$+$\Delta \theta_{i,k}$; 
        \State
        $M_{DH(i,k)}$ = Get\_Matrix($A$, $B$, $C$, $D$);
        \textbf{See Eq.1} 
    \EndFor
    \end{algorithmic}
        
    \begin{algorithmic}
    \For{$k$ \textbf{in} $N_{Dof(i)}$ - 1}
        \State
        $M_{DH(i,k+1)}$ = Update\_MDH($M_{DH(i,k)}$, $M_{DH(i,k+1)}$); \textbf{See Eq.2} 
    \EndFor
    \end{algorithmic}
\end{algorithmic}

\begin{algorithmic}
\For{$i$ \textbf{in} $N_{branch}$}
    \For{$k$ \textbf{in} $N_{Dof(i)}$}
        \State
        $x_{i,k}$ = $M_{DH(i,k,0,3)}$; 
        $y_{i,k}$ = $M_{DH(i,k,1,3)}$;
        $z_{i,k}$ = $M_{DH(i,k,2,3)}$;
        \State
        $P_{new(i,k)}$ = $R_x$ $R_y$ $R_z$($x_{i,k}$, $y_{i,k}$, $z_{i,k}$) + ($T_x$, $T_y$, $T_z$);\\
    \EndFor
\EndFor
\textbf{return $P_{new}$}
\end{algorithmic}
\end{algorithm}

\textbf{Human kinematics model based on DH parameters.}
DH parameter~\cite{craig2009introduction} is a method to describe the coordinate system of connecting links. 
The schematic diagram of the DH parameter can be seen in the right part of Fig.~\ref{fig:right hand}, where $a$ is the link length, $d$ is the link offset, $\alpha$ is the twist angle, $\theta$ is the joint angle.
These four parameters are DH parameters. 
Each degree of freedom (DOF) has a set of DH parameters.
We use DH parameters to establish the human kinematics model as shown in Fig.~\ref{fig:dh_para_model}. 
Some parameters of the model are fixed, which determines the connection relationship between bones, while others determines the rotation relationship between bones and the length of bones.
In the DH parameter table in Fig.~\ref{fig:dh_para_model}, those marked with red triangles are variable parameters, and others are preset fixed parameters.
See Alg.~\ref{alg1} for the process of building a human kinematics model (DH parameter model) with DH parameters.
$\Delta a$, $\Delta d$, $\Delta \alpha$, $\Delta \theta$ are the change in DH parameters.
$R_x$, $R_y$, $R_z$ are the global rotation parameters.
$T_x$, $T_y$, $T_z$ are the global translation parameters.
In addition, they are the values output by the fully connected network.
Output $P_{new}$ is a new 3D pose. The value of $N_{branch}$ is 5, which is the number of DH branches.
$N_{Dof(i)}$ is the number of degrees of freedom (DOF) per branch.
First, the DH parameters are converted into the transformation matrix:
\begin{equation}
    \begin{aligned}
        M_{DH}=
        \begin{bmatrix}
          cos(\theta) & -sin(\theta) & 0 & a \\
          sin(\theta)cos(\alpha) & cos(\theta)cos(\alpha) & -sin(\alpha) & -dsin(\alpha) \\
          sin(\theta)sin(\alpha) & cos(\theta)sin(\alpha) & cos(\alpha) & dcos(\alpha) \\
          0 & 0 & 0 & 1
         \end{bmatrix}
         & 
    \end{aligned}
\end{equation}
where $a$ is the link length, $d$ is the link offset, $\alpha$ is the twist angle, $\theta$ is the joint angle. Next, The inner product is used to update the transformation matrix:
\begin{equation}
    M'_{DH(i,k+1)}=M_{DH(i,k)}M_{DH(i,k+1)}
\end{equation}
where $i$ is the index of the branch, and $k$ is the index of the degree of freedom in $branch_i$.
Then, a new 3D pose is extracted from $M_{DH}$. 
Finally, we globally rotate and translate the new 3D pose. 
Other details are illustrated in Fig.~\ref{fig:dh_para_model}.

\textbf{Constraints on DH parameter model}. We implemented two constraints on the DH parameter model.
\textbf{1)} We removed the redundant degrees of freedom (DOF). 
For example, we only set 1 DOF for the elbow and knee instead of 3, and the number of DOF is changed from 48 to 33 (the number of key points is 16). 
For details, see the human skeleton in Fig.~\ref{fig:dh_para_model}. 
This operation not only greatly reduces the parameters that the GAN needs to learn, but also prevents the generator from producing a human skeleton with unreasonable rotation direction.
\textbf{2)} We designed a DH parameter constraint table to limit the value of DH parameters.
We list the constraint table of the right-hand branch, as shown in Fig.~\ref{fig:right hand}.
The left side of Fig.~\ref{fig:right hand} is the constraint diagram of the elbow, and the middle side is the DH parameter constraint table of the whole right-hand branch. 
DH parameter constraint table is added to the last layer of fully connected network:
\begin{equation}
    P_{DH}=(1+\tanh(O_{FC}))\frac{T_{DH(max)}-T_{DH(min)}}{2}
\end{equation}
where $P_{DH}$ is the DH parameter, $O_{FC}$ is the output of the fully connected network,
$T_{DH}$ is the DH parameter constraint table (Fig.~\ref{fig:right hand}).

\begin{figure}[t]
\begin{center}
\includegraphics[width=0.95\linewidth]{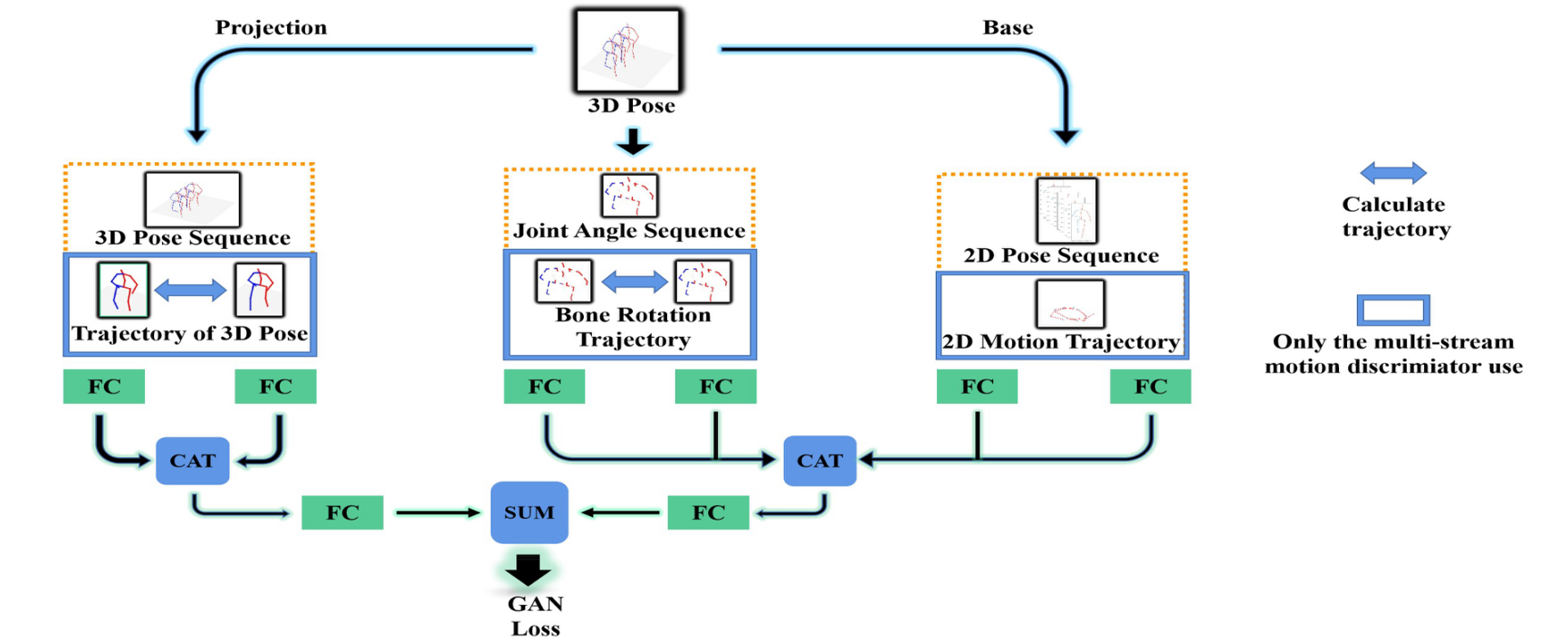}
\end{center}
\caption{\textbf{Multi-stream motion discriminator (MSMD).} It has 3 two-stream branches.
\textbf{Left}: 3D pose two-stream branch. 
\textbf{Middle}: Bone rotation two-stream branch.
\textbf{Right}: 2D pose two-stream branch. 
When we remove the branch in the blue box, it becomes a single-frame pose discriminator.}
   \label{fig:multi_stream}
\end{figure}

\subsection{Architecture}
Fig.~\ref{fig:framework_dh_aug} shows the overall framework of DH-AUG. 
Here we will describe various components of DH-AUG. 

\textbf{DH-generator.} We combine the DH parameter model and fully connected network (FC) to form a new generator called DH-generator.
The fully connected network first samples the 128-dimensional vector $z$ from the normal distribution as the input then generates DH parameters:  $\Delta a$, $\Delta d$, $\Delta \alpha$, $\Delta \theta$
(48 in total, includes bone length),
global rotation parameters: $R_x,R_y,R_z$, and global translation parameters: $T_x, T_y, T_z$.
These parameters are input into the DH parameter model.
After a series of operations described in Section 3.2, a new 3D pose will be generated. 
Finally, the 3D pose is projected~\cite{hartley2003multiple} to a 2D pose through camera parameters (from the original
dataset).
When DH-generator is used in single-frame pose estimation, only one 3D pose is generated by sampling one vector. However, when DH-generator is used in video pose estimation, a vector is sampled to generate a 3D pose sequence.

\textbf{Multi-stream motion discriminator (MSMD).}
To generate a skeleton video with motion continuity, we add timing information to the discriminator, as shown in Fig.~\ref{fig:multi_stream}). 
It has 3 two-stream branches.
1) 3D pose two-stream branch.
We input the 3D pose sequence and the trajectory of the 3D pose into this two-stream branch.
The trajectory of the 3D pose is calculated as follows:
\begin{equation}
    D_{3D}=\sum_{t=1}^{T}\sum_{i=0}^{I}(P_{3D_{(t,i)}}-P_{3D_{(t-1,i)}})
\end{equation}
where $P_{3D_{(t,i)}}$ is the 3D coordinate of the $i$th key point in frame $t$. 
2) Bone rotation two-stream branch.
We calculate the joint angle between adjacent bones, and the formula is as follows:
\begin{equation}
    A_{(t,i)}=\frac{V_{t,i} \cdot V_{t,i-1}}{L_{t,i}  L_{t,i-1}}
\end{equation}
where $V_{t,i}$ is the $i$th bone vector in frame $t$, $L_{t,i}$ is the $i$th bone length in frame $t$.
$i$ and $i-1$ are a pair of adjacent bones. 
Another input to this two-stream branch is the bone rotation trajectory:
\begin{equation}
    D_{Angle}=\sum_{t=1}^T\sum_{i=0}^{I}(A_{t,i}-A_{t-1,i})
\end{equation}
where $A_{t,i}$ is the $i$th joint angle in frame $t$. 
3) 2D pose two-stream branch.
We input the 2D pose sequence and 2D motion trajectory into this two-stream branch.
This branch mainly guides the generator to produce the correct viewpoint.
The calculation formula of 2D motion trajectory is as follows:
\begin{equation}
D_{2D}=\sum_{t=1}^{T}(P_{2D_{(t,root)}}-P_{2D_{(t-1,root)}})
\end{equation}
where $P_{2D_{(t,root)}}$ is the 2D coordinate of the root key point in frame $t$, $root$ represents the key point of the hip.

\textbf{Single-frame pose discriminator.} The single-frame pose discriminator we use is a simplified
version of the MSMD. Its structure is the content after removing the components in the blue box in Fig.~\ref{fig:multi_stream}.

\textbf{Training loss.} Loss used by our GAN is the objective function in improved Wasserstein GAN~\cite{gulrajani2017improved}.
The loss we finally use is as follows:
\begin{equation}
\gamma = \left\{
        \begin{aligned}
        1 &  & epoch >= \beta\\
        0 &  & epoch < \beta
        \end{aligned}
        \right.\\
\end{equation}
\begin{equation}
    \begin{aligned}
        \begin{aligned}
        L &= E[D_s(X_f)]-E[D_s(X_r)]+\alpha E[(\left\| \nabla_{\hat{X}} D_s(\hat{X}) \right\|_2-1)^2] \\
          &+ \gamma(E[D_m(X_f)]-E[D_m(X_r)]+\alpha E[(\left\| \nabla_{\hat{X}} D_m(\hat{X}) \right\|_2-1)^2])
        \end{aligned}
    \end{aligned}
\end{equation}
where $D_s$ represents the output of single-frame pose discriminator, $D_m$ represents the output of multi-stream motion discriminator, $\alpha$ represents the weight of gradient penalty,
$\gamma$ represents whether to turn on the multi-stream motion discriminator,
$X_f$ is fake data, $X_r$ is real data, $\hat{X}$ is randomly sampled data, $\beta$ is the epoch that turns on the multi-stream motion discriminator.
In our experiment, $\alpha$ is 10, $\beta$ is 4.

\textbf{Pose estimator.} 
In this paper, we use SemGCN~\cite{zhao2019semantic}, SimpleBaseline~\cite{martinez2017simple} and VPose~\cite{pavllo20193d} as single-frame 3D pose estimators, VPose~\cite{pavllo20193d} and PoseFormer~\cite{zheng20213d} as video 3D pose estimators, and Det~\cite{deepfakes}, CPN~\cite{chen2018cascaded}, HR~\cite{sun2019deep} and ground truth as 2D pose estimators.

\textbf{About the use of synthetic data.} 
Each epoch generates the same number of data pairs as the training set and packs them into a new dataset.
Then in the next epoch, we will train the 3D pose estimator on the new dataset and the original dataset.

\section{Experiments}
\subsection{Implementation Details}
We use the fully connected network as the backbone network. 
See supplementary material for the specific structure of generator and discriminator.
When pose augmentation is performed for the video pose estimator, we first train the single-frame pose discriminator for 4 epochs and then turn on the multi-stream motion discriminator.
Single-frame: batch size is 1024, video: batch size is 512.
The pose estimator uses the Adam optimizer with a learning rate of 1e-4, 1e-3, or 2e-3.
The first 50 epochs use linear attenuation, and the subsequent epochs attenuate each epoch by 5\% to 10\%.
Both generator and discriminator use Adam optimizer, and the learning rate remains 1e-4 unchanged.
The training is carried out on one 1080ti GPU. 
Training about 100 to 140 epochs. 
The data used to train 2D-3D pose lifting network and DH-AUG are consistent. For example, In the weakly-supervised settings, both the pose lifting network and DH-AUG are trained using S1 in H36M.
See supplementary material for DH parameter constraint table, model structure, etc.

\subsection{Datasets}
\textbf{Human3.6M}~\cite{ionescu2014human3} is the largest benchmark dataset. Subjects 1, 5, 6, 7, 8 are used as the training set, and subjects 9, 11 are used as the test set. 
In case of weak supervision, S1 or S1, S5 shall be used for training, and S9, S11 shall be used for evaluation. 
MPJPE was used as evaluation criteria.

\textbf{MPI-INF-3DHP}~\cite{mehta2017vnect} and \textbf{3DPW}~\cite{von2018recovering} are large 3D datasets containing complex outdoor scenes.
Instead of using them for training, we use their test sets to evaluate the model’s generalization ability to unseen environments.
Evaluation criteria: PCK, AUC, MPJPE (MPI) and PA-MPJPE (3DPW).

\textbf{LSP}~\cite{johnson2010clustered} and \textbf{MPII}~\cite{andriluka20142d} are two 2D pose datasets containing a large number of outdoor scenes.
We selected several difficult pictures for qualitative experiments.

\textbf{DH-3DP:} We synthesized a dataset with more than 1 million 2D-3D data pairs using DH-AUG. The synthesis method of this dataset is: S15678 of H36M is used as the training set to train DH-AUG, with a total of 110 epochs. We use the pretrained DH-AUG to generate more than 1 million 2D-3D data pairs. See the supplementary materials for more details.

\begin{table}[!t]
\scriptsize
\centering
\caption{{\textbf{Results of using DH-AUG in video 3D pose estimation}. f represents the number of input frames. The evaluation criteria uses 
MPJPE. We downsample the frames used by a factor of 10. We use VPose~\cite{pavllo20193d} and Poseformer~\cite{zheng20213d} as 3D pose estimators. And DET~\cite{deepfakes}, CPN~\cite{chen2018cascaded}, HR~\cite{sun2019deep} and GT are used as 2D pose estimators. (It is worth noting that PoseAug~\cite{gong2021poseaug} is designed for single-frame pose estimator.)}
}
\newcommand{\TableEntry}[2]{\textbf{#1}~\scriptsize{\green{(-#2)}}}

\begin{tabular}{l|cccc|cccc}
\multicolumn{1}{c|}{} & \multicolumn{4}{c|}{MPI-3DHP~($\downarrow$)}  & \multicolumn{4}{c}{H36M~($\downarrow$)} \\ 
\hline
Method & \multicolumn{1}{c}{DET} & \multicolumn{1}{c}{CPN} & \multicolumn{1}{c}{HR} & \multicolumn{1}{c|}{GT} &  \multicolumn{1}{c}{DET} & \multicolumn{1}{c}{CPN} & \multicolumn{1}{c}{HR} & \multicolumn{1}{c}{GT}\\ 
\hline

VPose~\cite{pavllo20193d} (f=9) & 97.56 & 94.26 & 90.83 & 90.7 & 61.47 & 55.74 & 53.79 & 42.14 \\  

\textbf{Vpose+DH-AUG (f=9)} & \textbf{84.23} & \textbf{84.76} & \textbf{82.57} & \textbf{80.39} & \textbf{60.81} & \textbf{55.66} & \textbf{53.04} & \textbf{41.21} \\
\hline

VPose~\cite{pavllo20193d} (f=27) & 101.99 & 97.33 & 94.62 & 91.76 & 61.84 & 56.57 & 52.89 & 42.18\\

\textbf{Vpose + DH-AUG (f=27)} & \textbf{86.34} & \textbf{88.38} & \textbf{84.37} & \textbf{80.85} & \textbf{61.19} & \textbf{56.07} & \textbf{52.57} & \textbf{41.52} \\
\hline

PoseFormer~\cite{zheng20213d} (f=9) & 95.09 & 88.01 & 82.38 & 85.28 & 63.28 & 56.47 & 54.24 & 42.02 \\

\textbf{PoseFormer + DH-AUG (f=9)} & \textbf{81.99} & \textbf{81.13} & \textbf{76.07} & \textbf{76.25} & \textbf{63.13} & \textbf{55.73} & \textbf{53.32} & \textbf{39.29}  \\
\hline

PoseFormer~\cite{zheng20213d} (f=27) & 92.71 & 86.38 & 83.16 & 84.67 & 62.26 & 55.00 & 53.34 & 39.63\\

\textbf{PoseFormer + DH-AUG (f=27)} & \textbf{81.04} & \textbf{77.13} & \textbf{72.18} & \textbf{75.36} & \textbf{62.26} & \textbf{54.95} & \textbf{52.46} & \textbf{37.92}\\

\end{tabular}

\label{tab:video}
\end{table}

\begin{table}[t]
	\small
	\centering
	\caption{\small \textbf{Results on H36M and MPI.} Evaluation criteria: MPJPE. Best in bold. 
    }
    
    \newcommand{\TableEntry}[2]{{#1}~\scriptsize{\green{(-#2)}}}

	\label{tab:ablation-discriminator}
	\begin{tabular}{l|cc}
		Method & 3DHP~($\downarrow$) &  H36M~($\downarrow$)\\
		\hline 

		VPose~\cite{pavllo20193d} (f=27) & 91.76 & 42.18 \\

		Liu et al~\cite{liu2020attention} (f=243) & 91.86 &  42.70 \\

		Anatomy~\cite{chen2021anatomy} (f=27)  & 86.01 & 39.98 \\

		PoseFormer~\cite{zheng20213d} (f=27)  & 84.67 & 39.63 \\
        \hline
        PoseFormer (f=27)  + \textbf{DH-AUG (Ours)} & \textbf{75.36} & \textbf{37.92} \\
	\end{tabular}
	
\label{tab:video_sota}
\end{table}

\subsection{Pose Augmentation in Video Pose Estimation}
We use VPose~\cite{pavllo20193d} and PoseFormer~\cite{zheng20213d} as the 3D pose estimators and Det~\cite{deepfakes}, CPN~\cite{chen2018cascaded},
HR~\cite{sun2019deep}, and ground truth as the 2D pose estimators. Experiments were carried out with 9 and 27 frames.
Because H36M is large, we choose to use 10 times of downsampling data for training.
The results are shown in Table.~\ref{tab:video}.
It can be seen that DH-AUG can greatly increase the generalization ability of the video pose estimator.
It is worth noting that PoseAug~\cite{gong2021poseaug} is designed for a single-frame pose estimator.
It can not be directly used in a video pose estimator.
Table.~\ref{tab:video_sota} is the result on H36M and MPI.
It can be seen that our method outperforms other SOTA methods.

\begin{table}[b]
	\small
	\centering
	\caption{\textbf{Results on H36M (fully supervised)}. Evaluation criteria: MPJPE. Best in bold. * denotes the SOTA pose augmentation method.}
	\begin{tabular}{l|c}
		Method &  MPJPE~($\downarrow$)  \\
		\hline
		SemGCN (CVPR'19)~\cite{zhao2019semantic}  & 57.60 \\		
		
		Sharma (CVPR'19)\cite{sharma2019monocular}   & 58.00 \\

		Moon (ICCV'19)~\cite{moon2019camera}  & 54.40 \\

	    VPose (CVPR'19)~\cite{pavllo20193d}   & 52.70 \\
	
		*Li (CVPR'20)~\cite{li2020cascaded} & 50.90 \\
	
		*VPose + PoseAug (CVPR'21)~\cite{gong2021poseaug} & 50.20 \\ 	
		\hline

		\textbf{VPose + DH-AUG}   & \textbf{49.81}  \\	
	\end{tabular}
	\label{tab:h36m_fully}

\end{table}

\subsection{Pose Augmentation in Single-Frame Pose Estimation}
To be consistent with other methods, we use HR~\cite{sun2019deep} as the 2D pose estimator and VPose~\cite{pavllo20193d} as the 3D pose estimator. 
Table.~\ref{tab:h36m_fully} is the result on H36M.
It can be seen that our method outperforms other SOTA methods (fully-supervised).

To evaluate the model’s generalization ability, we only use H36M for training and use MPI and 3DPW as test sets.
Moreover, we use ground truth as 2D data and VPose~\cite{pavllo20193d} as the 3D pose estimator.
See Table.~\ref{tab:3dhp} for MPI test results.
See the right part of Table.~\ref{tab:single_frame} for 3DPW test results.
We can observe that our method achieves the best performance under all the metrics.

\begin{table}[t]
	\scriptsize
	\centering
	
	\caption{\textbf{Results on MPI (fully supervised)}. The evaluation criteria were PCK, AUC and MPJPE. CE means evaluation across datasets. Best in bold. * represents SOTA pose augmentation method. \textbf{S1 + S5}: Use S1 and S5 for training. }
	
	\begin{tabular}{l|c|ccc}
		Method & CE & MPJPE~($\downarrow$) & PCK~($\uparrow$) & AUC~($\uparrow$)  \\     
		\hline

		Mehta\cite{mehta2017monocular} & & 117.60 & 76.50 & 40.80  \\

		VNect~\cite{mehta2017vnect}  &  & 124.70 & 76.60 & 40.40 \\

		Multi Person~\cite{chu2017multi} & & 122.20 & 75.20 & 37.80  \\
	
		OriNet~\cite{luo2018orinet} & & {89.40} & 81.80 & 45.20 \\
		\hline
		
		LCN~\cite{ci2019optimizing} & \checkmark & - & 74.00 & 36.70 \\

		HMR~\cite{kanazawa2018end} & \checkmark & 113.20 & 77.10 &	40.70  \\

		SRNet~\cite{zeng2020srnet} & \checkmark & - & 77.60 & 43.80 \\

		RepNet~\cite{wandt2019repnet} & \checkmark  & 92.50 & 81.80 & 54.80\\

	    *Li~\cite{li2020cascaded} & \checkmark & 99.70 & 81.20 & 46.10\\
	
		VPose~\cite{pavllo20193d} & \checkmark  & 86.60 & - & -\\
				
		*VPose+PoseAug~\cite{gong2021poseaug}  & \checkmark & 73.00 & 88.60 & 57.30 \\
		\hline
		
		VPose+DH-AUG (S1+S5) & \checkmark & {72.93} &  {88.60} & {57.65}\\

		\textbf{VPose+DH-AUG} & \checkmark & \textbf{71.17} &  \textbf{89.45} & \textbf{57.93} \\
	\end{tabular}

	\label{tab:3dhp}
\end{table}

\begin{table}[t]
	\small
	\centering
	\caption{\textbf{Results on H36M and MPI (weakly supervised)}. Evaluation criteria:MPJPE. Best in bold.}
	
	\begin{tabular}{l|c|c|c|c}

		Train Set & \multicolumn{2}{c}{S1} & \multicolumn{2}{|c}{S1 + S5} \\
		\hline

		Method &  MPI & H36M & MPI & H36M  \\
		\hline

		VPose~\cite{pavllo20193d} & 116.4 & 65.2 & 93.5 & 57.9 \\		

		VPose+PoseAug~\cite{gong2021poseaug} & 90.3 & 56.7 & 77.9 & 51.3 \\	
		\hline

		\textbf{VPose+DH-AUG}   & \textbf{86.72} & \textbf{52.15} & \textbf{72.93} & \textbf{46.99} \\	

	\end{tabular}
	
	\label{tab:h36m_weakly}

\end{table} 

The effect of our method is more obvious when it is weakly-supervised.
Consistent with other pose augmentation methods, we used S1 or S1, S5 in H36M dataset for training and evaluated on H36M and MPI.
In addition, we use ground truth as 2D data and VPose~\cite{zheng20213d} as the 3D pose estimator.
The results are shown in Table.~\ref{tab:h36m_weakly}. 
It can be seen that DH-AUG outperforms the previous best method.

To further prove the generality of our method. 
We use SemGCN~\cite{zhao2019semantic}, SimpleBaseline~\cite{martinez2017simple} and, VPose~\cite{pavllo20193d} as the 3D pose estimator and Det~\cite{deepfakes}, CPN~\cite{chen2018cascaded}, HR~\cite{sun2019deep}, and 
ground truth as the 2D pose estimator.
The results are shown in Table.~\ref{tab:single_frame}.
It can be seen that our method outperforms the previous best pose augmentation method.

\begin{table}[!t]
\scriptsize
\centering
\caption{{\textbf{Results on H36M, MPI and 3DPW}. Different 2D and 3D pose estimators were used to evaluate the results before and after using 
DH-AUG. Consistent with previous experiments, DET~\cite{deepfakes}, CPN~\cite{chen2018cascaded}, HR~\cite{sun2019deep} and GT are used as 2D pose estimators, SemGCN~\cite{zhao2019semantic}, 
SimpleBaseline~\cite{martinez2017simple} and VideoPose~\cite{pavllo20193d} are used as 3D pose estimators, and +PoseAug denotes the result of the recent SOTA pose augmentation method~\cite{gong2021poseaug}. The evaluation criteria is MPJPE (MPI, H36M) and PA-MPJPE (3DPW).} 
}

\newcommand{\TableEntry}[2]{\textbf{#1}~\scriptsize{\green{(-#2)}}}
\begin{tabular}{l|cccc|cccc|c}
\multicolumn{1}{c|}{} & \multicolumn{4}{c|}{MPI-3DHP~($\downarrow$)}  & \multicolumn{4}{c|}{H36M~($\downarrow$)}  & \multicolumn{1}{c}{3DPW~($\downarrow$)} \\ 
\hline
Method & \multicolumn{1}{c}{DET} & \multicolumn{1}{c}{CPN} & \multicolumn{1}{c}{HR} & \multicolumn{1}{c|}{GT} &  \multicolumn{1}{c}{DET} & \multicolumn{1}{c}{CPN} & \multicolumn{1}{c}{HR} & \multicolumn{1}{c|}{GT}& \multicolumn{1}{c}{GT}\\ 
\hline
SemGCN~\cite{zhao2019semantic}  & 101.90 & 98.70 & 95.60 & 97.40 & 67.50 & 64.70 & 57.50 & 44.40 & 102.00\\
SemGCN + PoseAug~\cite{gong2021poseaug}  & 89.90 & 89.30  & 89.10  & 86.10 & 65.20  & 60.00 & 55.00 & 41.50 & 82.20\\
\textbf{SemGCN + DH-AUG} & \textbf{79.68} & \textbf{76.67} & \textbf{72.99} & \textbf{71.31} & \textbf{63.16} & \textbf{56.93} & \textbf{54.04} & \textbf{40.00} & \textbf{79.07} \\
\hline

SimpleBaseline~\cite{martinez2017simple}  & 91.10 & 88.80 & 86.40 & 85.30 & 60.50 & 55.60 & 53.00 & 43.30 & 89.40 \\

SimpleBaseline + PoseAug~\cite{gong2021poseaug} & 78.70 & 78.70 & 76.40 & 76.20 & 58.00 & 53.40 & 51.30 & 39.40 & \textbf{78.10} \\
\textbf{SimpleBaseline + DH-AUG} & \textbf{77.99} & \textbf{75.87} & \textbf{72.97} & \textbf{72.28} & \textbf{57.86} & \textbf{53.13} & \textbf{50.06} & \textbf{38.89} & {80.52} \\
\hline

VPose~\cite{pavllo20193d} (1-frame) & 92.60  & 89.80 & 85.60 & 86.60 & 60.00 & 55.20 & 52.70 & 41.80  & 94.60\\

VPose + PoseAug~\cite{gong2021poseaug} & 78.30 & 78.40 & 73.20 & 73.00 & 57.80 & 52.90 & 50.20 & 38.20 & 81.60\\
\textbf{VPose + DH-AUG (1-frame)} & \textbf{76.70}  & \textbf{74.82} & \textbf{71.07} & \textbf{71.17} & \textbf{57.66} & \textbf{52.52} & \textbf{49.81} & \textbf{37.01}  & \textbf{79.28} \\
\end{tabular}

\label{tab:single_frame}
\end{table}

\begin{figure}[t]
\begin{center}
\includegraphics[width=0.95\linewidth]{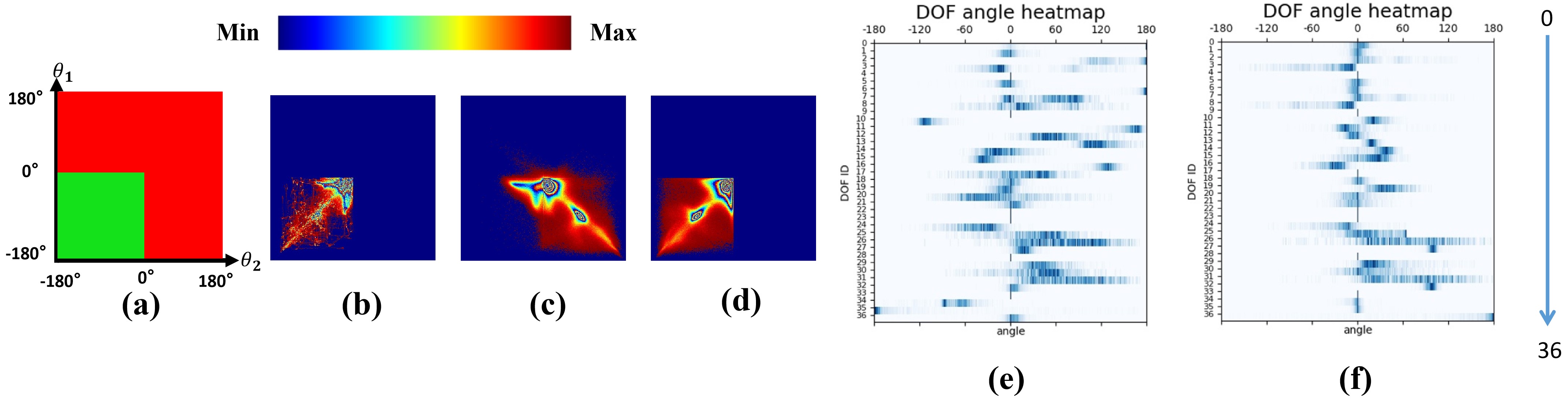}
\end{center}

\caption{\textbf{Data distribution. (b), (c), (d) is the data distribution of left knee and right knee joint angles. The normal rotation range of the knee is -180° to 0°. (e), (f) is the data distribution of all joint angles.} 
The amount of data in the (b), (c), (d) is the same. 
\textbf{(a)}: The green area is the normal area. 
The red area is the area where ambiguity occurs.
$\boldsymbol{\theta_1}$: Left knee joint angle. 
$\boldsymbol{\theta_2}$: Right knee joint angle. 
\textbf{(b)}: Data distribution of H36M datasets (before pose augmentation). 
\textbf{(c)}: Pose augmentation (no constraints). 
\textbf{(d)}: Pose augmentation (add constraints).
\textbf{(e)}: Pose augmentation (no constraints).
\textbf{(f)}: Pose augmentation (add constraints).
}
\label{fig:distribution}
\end{figure}

\textbf{Analysis of the data distribution.} To further verify the diversity of the data we generated, we visualized the data distribution of the left knee and right knee. Distribution of H36M (before augmentation) form a small and concentrated cluster, also showing a limited diversity (Fig.~\ref{fig:distribution} (b)). However, our method (DH-AUG) obtains a huge and decentralized cluster as shown in Fig.~\ref{fig:distribution} (d). This shows that DH-AUG generates more diverse pose, and also proves why our method can greatly enhance the generalization ability. The comparison of distribution before and after adding constraints will be introduced in section 4.6. In addition, we provide the distribution of all joint angles, as shown in Fig. \ref{fig:distribution} (e) and (f).

\begin{figure}[t]
\begin{center}
\includegraphics[width=0.95\linewidth]{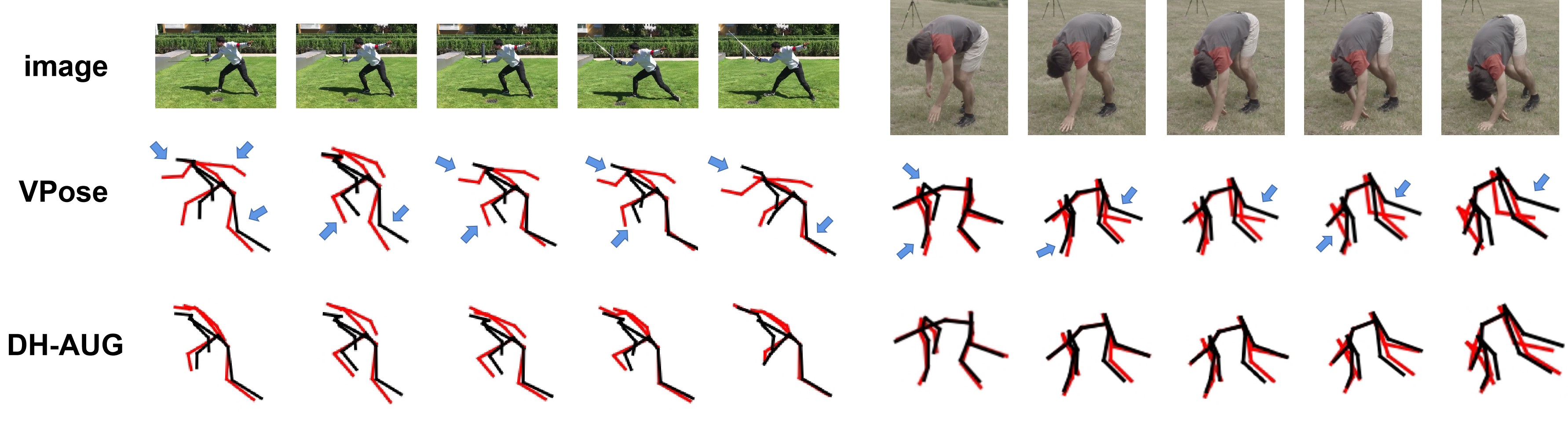}
\end{center}

\caption{\textbf{Qualitative results on MPI and 3DPW.} The black pose is the ground truth. The blue arrow points to the location of the main correction. \textbf{More qualitative results are shown in the supplementary material.}}

   \label{fig:qualitative_result}
\end{figure}

\begin{table}[t]
	\small
	\centering

	\caption{\small \textbf{Ablation study.} \textbf{DHG}: DH-generator.
    \textbf{BR}: Bone rotate two-stream module. \textbf{2DP}: 2D Pose two-stream.
    \textbf{DHT}: DH parameter constraint table.
    }
    
    \newcommand{\TableEntry}[2]{{#1}~\scriptsize{\green{(-#2)}}}

	\label{tab:ablation-discriminator}
	\begin{tabular}{l|cc}
		Method & 3DHP~($\downarrow$) &  H36M~($\downarrow$)\\
		\hline 

		Baseline & 90.70 & 42.14 \\

		+ DHG + BR & 88.97 & 43.65 \\

		+ DHG + BR + 2DP & 84.12 & 41.31 \\

		+ DHG + BR + 2DP + 3DP & 82.86 & 41.20 \\

		+ DHG + BR + 2DP + 3DP + DHT & 80.39 & 41.21 \\

	\end{tabular}
	
\label{tab:ablation}
\end{table} 

\begin{figure}[t]
\begin{center}
\includegraphics[width=0.95\linewidth]{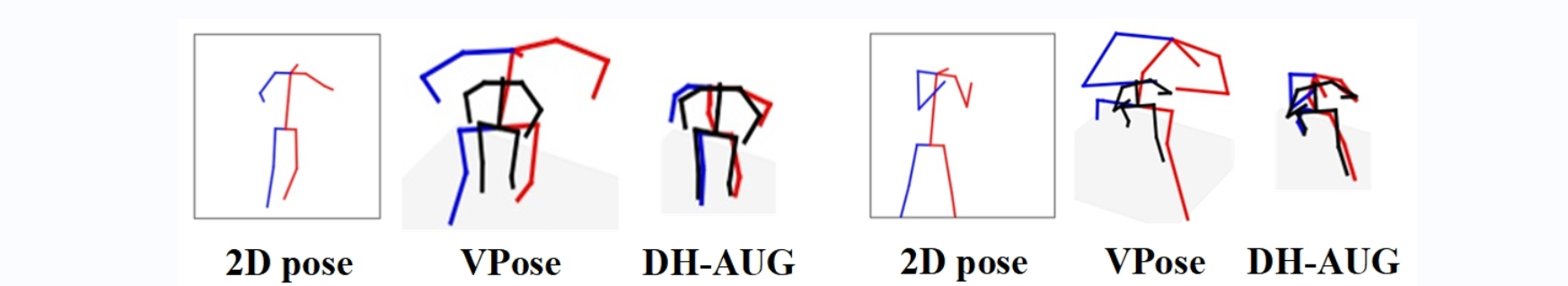}
\end{center}

\caption{\textbf{Scale problem.} Columns 1, 4 are 2D poses, columns 2, 5 are the results before pose augmentation,
and columns 3, 6 are the results of using DH-AUG. The black pose is the ground truth.}
   \label{fig:scale_problem}
\end{figure}

\begin{figure}[t]
\begin{center}
\includegraphics[width=0.95\linewidth]{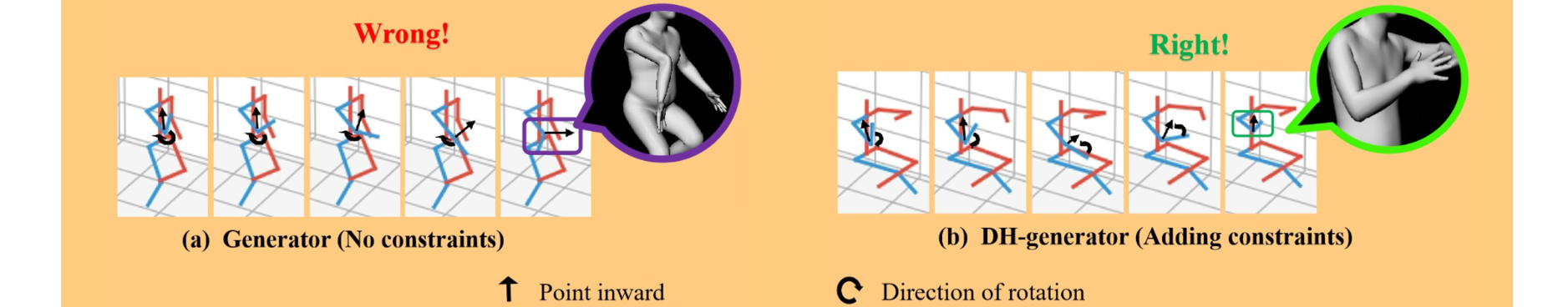}
\end{center}

\caption{\textbf{Skeleton video generated by DH-AUG.} \textbf{(a)} No constraints. \textbf{(b)} Adding constraints. SMPL is only used for visualization. \textbf{See the supplementary material for more skeleton videos.}}
\label{fig:skeleton video}
\end{figure}

\subsection{Qualitative Results}
We select difficult figures from several datasets for estimation, as shown in Fig.~\ref{fig:qualitative_result}.
The pose estimator enhanced with DH-AUG can get results with more correct action, better scale matching, and higher accuracy. \textbf{More qualitative results are shown in the supplementary material.}

We selected 2 frames of data close to the camera and found that DH-AUG can solve the scale problem shown in Fig.~\ref{fig:scale_problem}.
The reason for the scale mismatch is that the human motion in H36M is concentrated in one range,
which makes the model unable to fully learn the relationship between bone length and distance.
\subsection{Ablation Study}
\textbf{BR, 2DP, 3DP.}  
\textbf{BR} is the bone rotation two-stream branch, which is used to constrain the joint angle parameters produced by the generator.
\textbf{2DP} is the 2D pose two-stream branch,
which is mainly used to constrain global translation parameters and motion trajectories.
\textbf{3DP} is the 3D pose two-stream branch, which is mainly used to constrain global
rotation parameters and enable the model to learn bone length information. 

\textbf{Effect of DH parameter constraint table.}
\textbf{DHT} is the DH parameter constraint table. Fig.~\ref{fig:distribution} is the data distribution of the left knee and right knee. The normal rotation range is about -180° to 0°. Before adding the DHT (Fig.~\ref{fig:distribution} (c)), the data distribution is asymmetric and unreasonable. Fig.~\ref{fig:distribution} (c) has a lot of outward rotation values (between 0° and 180°, the knee cannot be external rotation), which indicates that the generator has learned the wrong human kinematics information. This causes the generator to produce the skeleton video shown in Fig.~\ref{fig:skeleton video} (a). However, by observing Fig.~\ref{fig:distribution} (d), it will be found that the distribution is symmetrical and reasonable. After adding constraints, the skeleton video generated by DH-AUG is shown in Fig.~\ref{fig:skeleton video} (b). 

By observing the table.~\ref{tab:ablation}, we can see that the performance is improved by gradually adding \textbf{BR},
\textbf{2DP}, \textbf{3DP}, and \textbf{DHT} modules, which verifies the effectiveness of these modules.

\subsection{Limitation Analysis}
Although our method increases the generalization ability of 3D human pose estimation, our method still has some limitations. The DH parameter constraint table we use is manually set according to personal experience. This increases the number of hyper-parameters that need to be adjusted. Although it will not have a great impact on the final result, it increases some workload.

\section{Conclusion}
In this paper, we propose a pose augmentation solution, which we call DH-AUG.
DH-AUG has a special kinematics model called the DH parameter model, which weakens the angle ambiguity (multiple solutions).
We use 3 common single-frame 3D pose estimators and 2 video 3D pose estimators to experiment. 
Extensive experiments demonstrate that DH-AUG can greatly increase the generalization ability of the pose estimator.

\noindent\textbf{Acknowledgements.}
This work was supported in part by the National Natural Science Foundation of China under Grant 62192784 and Grant 61871052. 

\clearpage
%

\bibliographystyle{splncs04}
\bibliography{egbib}

\end{document}